\documentclass{article}

\usepackage{enumitem}

\usepackage{microtype}
\usepackage{graphicx}
\usepackage{subcaption}
\usepackage{booktabs} 

\usepackage{hyperref}

\usepackage[preprint]{icml2026}

\usepackage{amsmath}
\usepackage{amssymb}
\usepackage{mathtools}
\usepackage{amsthm}

\usepackage[capitalize,noabbrev]{cleveref}

\usepackage{multirow}
\usepackage[table,xcdraw]{xcolor}
\usepackage[normalem]{ulem}
\useunder{\uline}{\ul}{}
\usepackage{pifont}
\usepackage{tabularx}
\newcolumntype{Y}{>{\centering\arraybackslash}X}
\usepackage{siunitx}

\theoremstyle{plain}

\theoremstyle{definition}

\theoremstyle{remark}

\usepackage[textsize=tiny]{todonotes}

\icmltitlerunning{GRACE: Adaptive Backbone Scaling for Class Incremental Learning}

\begin{document}

\twocolumn[
  \icmltitle{Grow, Assess, Compress: Adaptive Backbone Scaling for Memory-Efficient Class Incremental Learning}



  \icmlsetsymbol{equal}{*}

  \begin{icmlauthorlist}
    \icmlauthor{Adrian Garcia-Castañeda}{comp,sch}
    \icmlauthor{Jon Irureta}{comp,ehu}
    \icmlauthor{Jon Imaz}{comp}
    \icmlauthor{Aizea Lojo}{comp}
  \end{icmlauthorlist}

  \icmlaffiliation{comp}{Ikerlan Technology Research Center, Arrasate, Spain}
  \icmlaffiliation{sch}{Goi Eskola Politeknikoa, Mondragon Unibertsitatea, Arrasate, Spain}
  \icmlaffiliation{ehu}{University of the Basque Country UPV/EHU, Spain}

  \icmlcorrespondingauthor{Adrian Garcia-Castañeda}{adrian.garcia@ikerlan.es}

  \icmlkeywords{Machine Learning, ICML}

  \vskip 0.3in
]



\printAffiliationsAndNotice{}  

\begin{abstract}
  Class Incremental Learning (CIL) poses a fundamental challenge: maintaining a balance between the plasticity required to learn new tasks and the stability needed to prevent catastrophic forgetting. While expansion-based methods effectively mitigate forgetting by adding task-specific parameters, they suffer from uncontrolled architectural growth and memory overhead. In this paper, we propose a novel dynamic scaling framework that adaptively manages model capacity through a cyclic "\textbf{GR}ow, \textbf{A}ssess, \textbf{C}ompr\textbf{E}ss" (GRACE) strategy. Crucially, we supplement backbone expansion with a novel saturation assessment phase that evaluates the utilization of the model's capacity. This assessment allows the framework to make informed decisions to either expand the architecture or compress the backbones into a streamlined representation, preventing parameter explosion. Experimental results demonstrate that our approach achieves state-of-the-art performance across multiple CIL benchmarks, while reducing memory footprint by up to a 73\% compared to purely expansionist models.
\end{abstract}

\begin{figure}[ht]
    \centering
    \begin{subfigure}[b]{0.5\columnwidth}
        \centering
        \includegraphics[width=\textwidth]{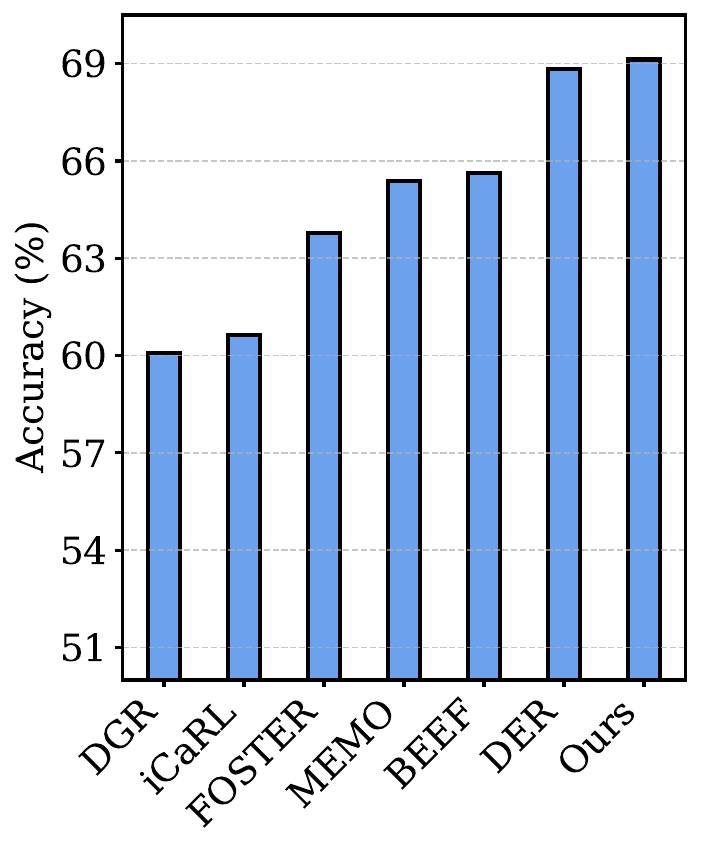}
    \end{subfigure}%
    \hfill
    \begin{subfigure}[b]{0.5\columnwidth}
        \centering
        \includegraphics[width=\textwidth]{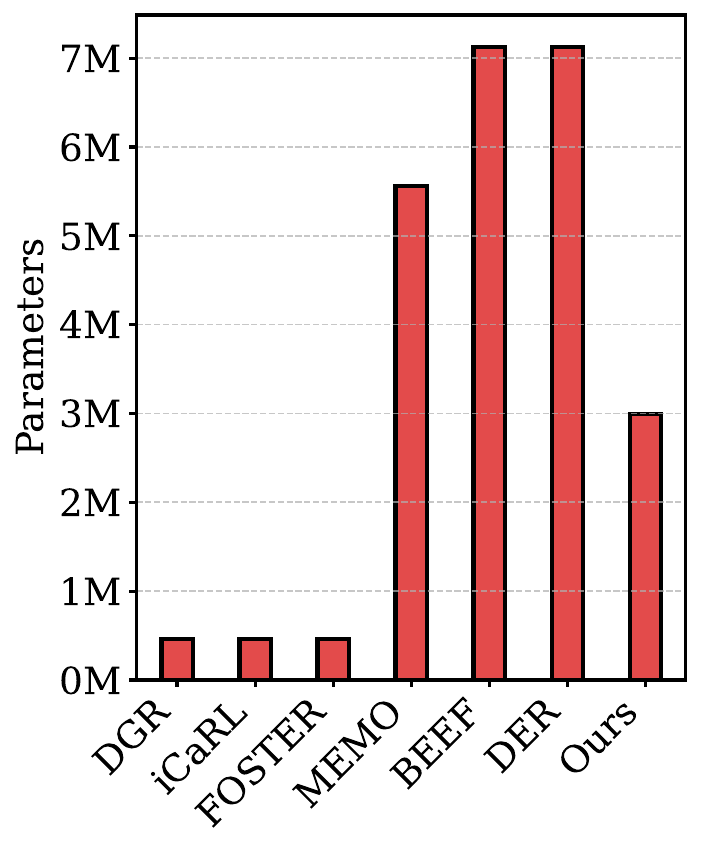}
    \end{subfigure}
    \begin{subfigure}[b]{\columnwidth}
        \centering
        \includegraphics[width=\textwidth]{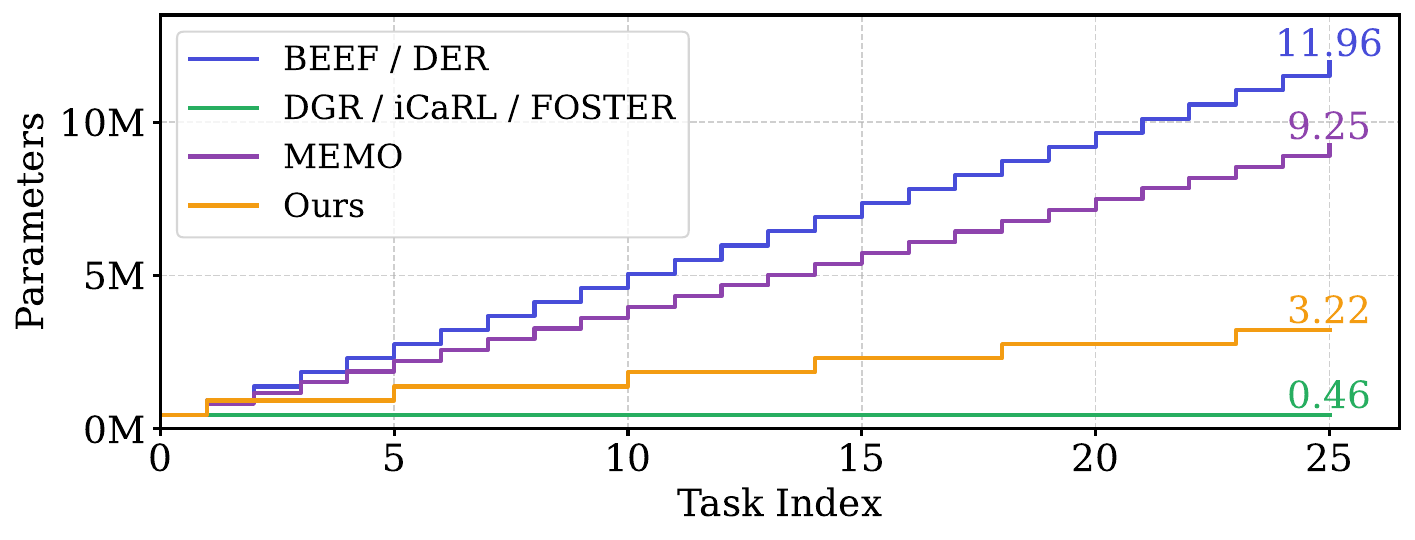}
    \end{subfigure}

    \caption{Mean average accuracy (top left) and parameter count (top right) on CIFAR-100 across different split protocols. The bottom panel illustrates the expansion dynamics for the \textit{Base 50 Inc 2} setting. Unlike baseline methods that expand arbitrarily each task or remain static, our proposed framework selectively grows the network only when necessary, obtaining state-of-the-art performance with a significantly lower parameter count.}
    \label{fig:intro}
\end{figure}

\section{Introduction}

The rapid evolution of deep learning has led to models of unprecedented scale and complexity \cite{brown2020gpt3}. However, the performance of these models is strictly tied to their training data, which inevitably becomes obsolete over time \cite{bayram2022conceptdrift}. While updating these models is essential, retraining from scratch is often infeasible due to the substantial costs of storing training data \cite{krempl2014storageconstraints}, high computational requirements \cite{strubell2019computationalreq}, and increasing legal constraints regarding data privacy \cite{europe2016gdpr}.

Continual learning addresses these challenges by enabling models to acquire new information sequentially without full retraining \cite{parisi2019clsurvey}. This paradigm requires a delicate balance between plasticity, which refers to the ability to integrate new knowledge, and stability, the preservation of legacy information; this trade-off is known as the \textit{stability-plasticity dilemma} \cite{grossberg2012spdilemma}. At the core of this difficulty lies catastrophic forgetting, a phenomenon where updating parameters for new tasks leads to the sudden erasure of previously learned representations \cite{mccloskey1989catastrophicforgetting}.

Within continual learning, Class Incremental Learning (CIL) represents a specific scenario where the evolving information is structured as a stream of previously unseen classes grouped in incremental tasks \cite{rebuffi2017icarl}. Among continual learning paradigms \cite{van2022threetypescl}, CIL is uniquely significant, as its inherent complexity makes it the primary benchmark for investigating the stability-plasticity dilemma \cite{wu2021cildilemma1, kim2023cildilemma2}.

In many real-world incremental scenarios, the total number of tasks and the composition of the data stream are unknown \textit{a priori} \cite{aljundi2019taskfree}. In these cases, static architectures often fail due to limited representational capacity \cite{knoblauch2020perfectmemory}. Consequently, expansion-based strategies have emerged as a dominant approach, isolating task-specific weights to prevent negative interference \cite{rusu2016pnn}. However, these methods introduce a new set of challenges: they often lead to unbounded parameter growth and the learning of redundant feature representations \cite{zhou2024cilsurvey}, making them unsustainable for resource-constrained environments, \textit{e.g.} the industrial sector.

To overcome these limitations, we introduce GRACE, a dynamic framework for continual learning that intelligently manages backbone scaling by regulating network capacity through three main contributions: \textbf{(i) Dynamic Capacity Management}, which utilizes a novel saturation-aware mechanism based on effective rank to make informed decisions on when to expand the architecture, thereby constraining the linear parameter growth of traditional expansion-based methods; \textbf{(ii) Enhanced Compression Phase}, that leverages an importance-aware student initialization, as well as both logit- and feature-level distillation to consolidate backbones with minimal performance impact; and \textbf{(iii) Resource-Aware Versatility}, providing adjustable configurations to tweak the expansion rate, which allow the framework to be adapted to diverse use cases, from memory-constrained edge devices to high-performance servers. By implementing the cyclic ``Grow, Assess, Compress'' mechanism, GRACE ensures a controlled architectural expansion that achieves state-of-the-art performance with a significant reduction in memory cost, translating to an advantage of more than 4 points in memory-aligned evaluations. To facilitate reproducibility, we release our source code at \url{https://github.com/ai-digilab/GRACE}.

The remainder of this paper is organized as follows: Section~\ref{sec:relatedwork} reviews related work in CIL. Section~\ref{sec:grace} describes the proposed GRACE framework. Section~\ref{sec:experiments} presents the experimental results and a comparison with state-of-the-art methods. Finally, Section~\ref{sec:conclusion} concludes the paper.

\section{Related Work}\label{sec:relatedwork}

\paragraph{Class Incremental Learning}

Class Incremental Learning (CIL) aims to learn a sequence of classes divided in incremental tasks, where only the data of the current task is available. Traditionally, methods have been categorized into: (a) \textbf{Regularization-based methods}, which impose constraints on weight updates to protect important parameters and mitigate negative interference; (b) \textbf{Replay-based methods}, which store a subset of past samples or employ generative models to produce synthetic exemplars to prevent forgetting; and (c) \textbf{Dynamic networks}, which adapt the model architecture to accommodate the evolving class stream. However, from a structural perspective, we can group these families into fixed-capacity methods (a and b), which optimize a static model, and dynamic networks, which evolve their topology over time.

\paragraph{Fixed-Capacity Methods}

Fixed-capacity methods typically employ replay buffers and regularization to mitigate catastrophic forgetting as the model incrementally learns novel concepts. EWC \cite{kirkpatrick2017ewc} estimates the Fisher Importance Matrix to constrain the update of parameters essential to previous tasks. iCaRL \cite{rebuffi2017icarl} combines inter-task knowledge distillation with a nearest-mean-of-exemplars classifier to reduce the inherent bias toward newly introduced classes. Recently, DGR \cite{he2024dgr} addresses dual imbalance in CIL (inter-task and intra-task) by utilizing gradient reweighting for unbiased optimization and distribution-aware knowledge distillation. While effective for small-scale problems, fixed-size approaches often struggle with the stability-plasticity trade-off as the number of tasks grows, eventually leading to performance degradation due to fixed representational capacity \cite{knoblauch2020perfectmemory}.

\begin{figure*}[ht]
  \vskip 0.1in
  \begin{center}
    \centerline{\includegraphics[width=0.98\textwidth]{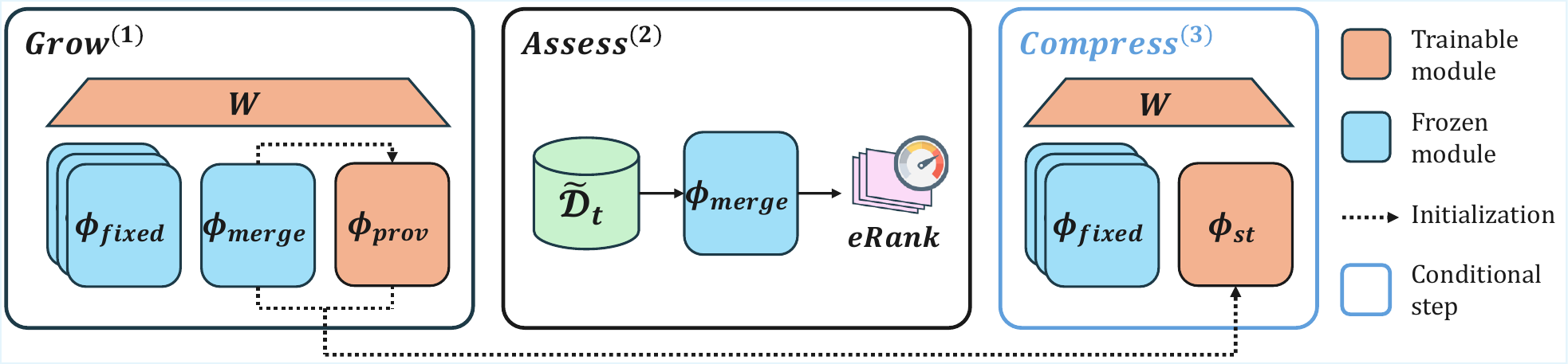}}
    \caption{
      Overview of the proposed GRACE framework (best viewed in colour). The incorporation of each novel task follows a three-stage pipeline: (1) \textit{Grow}, the architecture expansion phase (Section~\ref{sec:growphase}); (2) \textit{Assess}, the capacity evaluation phase (Section~\ref{sec:assessphase}); and (3) \textit{Compress}, the model consolidation phase (Section~\ref{sec:compressphase}).
    }
    \label{fig:graceframework}
    \vspace{-10pt}
  \end{center}
\end{figure*}

\paragraph{Dynamic Networks and Expansion-Based Models}

To avoid the limitations of static architectures, dynamic networks expand the number of parameters as new tasks arrive, thus isolating the network weights to mitigate interference among tasks. DEN \cite{yoon2017den} selectively retrains relevant neurons and expands the network only when a loss-based threshold is exceeded, utilizing group-sparsity and neuron duplication to mitigate forgetting. PNN \cite{rusu2016pnn} allocates a new architectural column per task and utilizes lateral connections for knowledge transfer, whereas P\&C \cite{schwarz2018pandc} introduces a ``Progress and Compress'' strategy that leverages an active column to learn new features before consolidating them into a static knowledge base, being EWC the content preservation method. AANets \cite{liu2021aanets} employ bi-level optimization to aggregate plastic and stable residual blocks, balancing new knowledge acquisition with legacy preservation.

Building on the multi-column philosophy presented by PNN, DER \cite{yan2021der} trains a dedicated backbone for each task, fusing concatenated features through a shared classifier. To avoid the resulting memory overhead, FOSTER \cite{wang2022foster} employs a gradient-boosting-inspired framework to expand new modules and subsequently consolidate them via knowledge distillation. Instead of expanding entire backbones, MEMO \cite{zhou2022memo} finds that shallow layers of different backbones are similar and generalizable; consequently, it focuses expansion exclusively on the deepest layers, which contain the most task-specific features. BEEF \cite{wang2022beef} introduces energy-based models to learn bi-compatible representations that allow for independent network growth. TagFex \cite{zheng2025tagfex} extends DER by incorporating a task-agnostic self-supervised module to enrich the expansion process with more generalizable features, while TCIL \cite{huang2023tcil} utilizes both logit- and feature-level distillation to allow inter-task knowledge transfer during the expansion phase. CREATE \cite{chen2025create} leverages per-class autoencoders and reconstruction error for classification, employing knowledge distillation to balance the continuous adaptation of the feature extractor with the preservation of past knowledge.

Departing from standard Convolutional Neural Networks, DyTox \cite{douillard2022dytox} leverages Vision Transformers (ViT) by learning task-specific tokens instead of full network replicas. Similarly, L2P \cite{wang2022l2p} and DualPrompt \cite{wang2022dualprompt} pioneered the use of prompt-based learning in this context. DNE \cite{hu2023dne} introduces network expansion in ViT with dense connections between the intermediate layers of task-expert networks.


\paragraph{Discussion}

Our proposal, GRACE, alternates between representation learning and structural consolidation. Drawing from DER, we utilize provisional modules to learn task-specific features without interfering with stable weights. To address the scalability issues inherent in DER, we incorporate a fusion mechanism inspired by FOSTER. While this behaviour mirrors the ``expand-and-compress'' cycle of FOSTER, our method diverges by maintaining conditional plasticity: unlike FOSTER, which strictly collapses the architecture back to a single backbone, our framework permits sustained growth if the representational capacity of the compressed network is insufficient to capture the complexity of new classes.

\section{Methodology}\label{sec:grace}

\subsection{Problem Formulation}

In the Class Incremental Learning (CIL) setting, a model is sequentially updated over a stream of $T$ tasks $\{\mathcal{D}_1, \mathcal{D}_2, \dots, \mathcal{D}_T\}$. Each task $t \in \{1, \dots, T\}$ is defined by a training dataset $\mathcal{D}_t = \{(x_i^t, y_i^t)\}_{i=1}^{n_t}$, where $x_i^t \in \mathcal{X}$ represents an input sample from the input space (\textit{e.g.}, images) and $y_i^t \in \mathcal{Y}_t$ is its corresponding ground-truth label. A fundamental constraint in CIL is that the label sets are disjoint across different tasks: $\mathcal{Y}_t \cap \mathcal{Y}_{t'} = \varnothing \quad \forall \; t \neq t'$.

At each task $t$, the objective is to expand the model's prediction space to encompass all categories encountered thus far. This cumulative label space is defined as $\tilde{\mathcal{Y}}_t = \bigcup_{i=1}^t \mathcal{Y}_i$. Consequently, the goal is to learn a mapping function $f: \mathcal{X} \to \tilde{\mathcal{Y}}_t$ that minimizes classification error across all observed classes in $\tilde{\mathcal{Y}}_t$ while maintaining a balance between the plasticity required to learn $\mathcal{Y}_t$ and the stability needed to preserve knowledge of $\tilde{\mathcal{Y}}_{t-1}$. We aim to tackle the CIL problem by balancing performance and memory constraints.

We adopt a rehearsal-based strategy by maintaining an exemplar buffer $\mathcal{E}$ with a fixed capacity $M$. At any given task $t > 1$, this buffer stores a representative subset of samples from previously encountered classes, such that $\mathcal{E} \subset \bigcup_{i=1}^{t-1} \mathcal{D}_i$ and $|\mathcal{E}| \leq M$. During the training phase of task $t$, the model is optimized using the expanded dataset $\tilde{\mathcal{D}}_t = \mathcal{D}_t \cup \mathcal{E}$.

\subsection{Methodology Overview}

To optimize the balance between representational plasticity and memory efficiency, GRACE (see Figure~\ref{fig:graceframework}) operates through a decision-driven cyclic process involving four different types of architectural components: 
\begin{itemize}
    \item A set of \textit{fixed backbones}, $\phi_{fixed}$, which have reached their \textit{maximum} capacity and will remain static during subsequent tasks.
    \item A single \textit{mergeable backbone}, $\phi_{merge}$, which is frozen during expansion but still has room for new knowledge.
    \item A \textit{provisional backbone}, $\phi_{prov}$, which is tasked with learning new representations at the expansion phase.
    \item A student backbone, $\phi_{st}$, which integrates the knowledge from both $\phi_{merge}$ and $\phi_{prov}$, and is optimized during the compression phase.
\end{itemize}

Building upon these components, the proposed framework operates through three interconnected stages (see Appendix~\ref{app:gracealgorithm} for the detailed algorithmic framework):
\begin{enumerate}[label=\Roman*.]
    \item \textbf{Grow:} Upon the arrival of a new task $t$, GRACE provisionally allocates a new trainable backbone, $\phi_{prov}$, to capture the novel features of $\mathcal{Y}_t$. During this phase, other backbones are frozen to prevent catastrophic forgetting of $\tilde{\mathcal{Y}}_{t-1}$.
    
    \item \textbf{Assess:} After training the provisional backbone, GRACE employs the normalized effective rank ($\tilde{eRank}$) to evaluate the saturation level of the feature space of the mergeable backbone, $\phi_{merge}$. This metric serves as a decision gate for architectural evolution:
    \begin{itemize}
        \item \textbf{Expansion Case:} If $\tilde{eRank}$ exceeds a saturation threshold (indicating capacity overload), the provisional backbone is promoted to the new mergeable backbone, while the former mergeable backbone is moved to the fixed knowledge base, $\phi_{fixed}$, where it remains immutable.
        \item \textbf{Consolidation Case:} If the threshold is not met, the framework identifies the expansion as redundant. The provisional backbone is then merged with the current mergeable backbone through the compression phase.
    \end{itemize}
    
    \item \textbf{Compress:} In cases where expansion is deemed unnecessary, GRACE merges the provisional and mergeable backbones into a single, optimized representation $\phi_{st}$. After the compression phase, $\phi_{st}$ becomes the new mergeable backbone for future tasks.
\end{enumerate}

By integrating these components, GRACE transforms the mapping function $f: \mathcal{X} \to \tilde{\mathcal{Y}}_t$ from a static or linearly growing architecture into a capacity-aware system that optimizes the balance between classification accuracy and memory footprint. The following sections elaborate on each phase within the GRACE framework.

\subsection{Architecture Expansion}\label{sec:growphase}

Following DER \cite{yan2021der}, we employ a dynamic architecture that instantiates a new feature extractor for each sequential task. To preserve previously acquired knowledge, all preceding extractors remain frozen during the expansion phase, with only the most recent backbone, $\phi_{prov}$, being trainable. The final prediction is generated by an expandable classifier $W$ that operates on the concatenated representation of all extractors:
\begin{equation}\label{eq:dermodel}
\begin{split}
    f_{exp}(x) = W^{\top} \bigl[ & \phi_{fixed}^0(x), \phi_{fixed}^1(x), \dots, \\
    & \phi_{merge}(x), \phi_{prov}(x) \bigr]
\end{split}
\end{equation}

To encourage the newly added encoder to capture discriminative features, we employ an auxiliary classifier, $W_{aux}$. This ensures that $\phi_{prov}$ not only distinguishes between classes within the current task but also maintains cross-task separability. This classifier operates on a label space of $|\mathcal{Y}_t| + 1$, encompassing the current task classes $\mathcal{Y}_t$ and a single unified category representing all preceding tasks.
\begin{equation}\label{eq:derauxiliary}
    f_{aux}(x) = W_{aux}^{\top}\phi_{prov}(x)
\end{equation}

The expansion phase is optimized via a cross-entropy loss applied to the main expanded model and the auxiliary head:
\begin{equation}\label{eq:derloss} 
    \mathcal{L}_{exp} = \ell_{CE}(f_{exp}(x), y) + \ell_{CE}(f_{aux}(x), y_{aux}),
\end{equation}
where $y_{aux}\in\{1,\dots,|\mathcal{Y}_t|+1\}$ denotes the ground truth targets in the auxiliary label space.

Finally, to correct the classifier bias toward new classes caused by the class imbalance in $\tilde{\mathcal{D}}_t$, the Weight-Align strategy \cite{zhao2020wa} is applied after optimizing the backbone and the expanded classifier.

\subsection{Capacity Evaluation and Selective Expansion}\label{sec:assessphase}

To decide whether the novel features introduced by the provisional backbone can be safely compressed with the mergeable backbone, the latter is subjected to a capacity evaluation test. This assessment is based on the normalized effective rank ($\tilde{eRank}$) of the features extracted by the mergeable backbone. The calculation of this saturation score and a detailed comparison with other measures can be found in Appendix~\ref{appendix:erank} and Appendix~\ref{appendix:erankcomparison}, respectively.

\paragraph{Threshold and Decay}

Having computed the $eRank$-based saturation score, we must define a threshold ($\tau_1$) to determine when a network expansion is required. While the effective rank increases steadily as the mergeable backbone incorporates more classes, each update slightly degrades the existing knowledge base. To account for this, we become progressively more inclined to maintain the expanded network after each consecutive compression.

We achieve the desired behaviour through a threshold decay factor, $\rho\in(0, 1]$, which reduces the base threshold over time. Every assessment phase that results in a decision to compress the provisional backbone decreases the threshold value. Conversely, if the decision is made to maintain the expanded network, the threshold resets to its initial value, $\tau_1$. This dynamic adjustment is formalized as follows:
\begin{equation}
    \tau_{t+1} = 
    \begin{cases} 
        \rho \cdot \tau_{t} & \text{if } \tilde{eRank} < \tau_t \\
        \tau_{1} & \text{if } \tilde{eRank} \geq \tau_t 
    \end{cases} 
\end{equation}

\subsection{Model Consolidation}\label{sec:compressphase}

\begin{figure}[t]
  \vskip 0.1in
  \begin{center}
    \centerline{\includegraphics[width=\columnwidth]{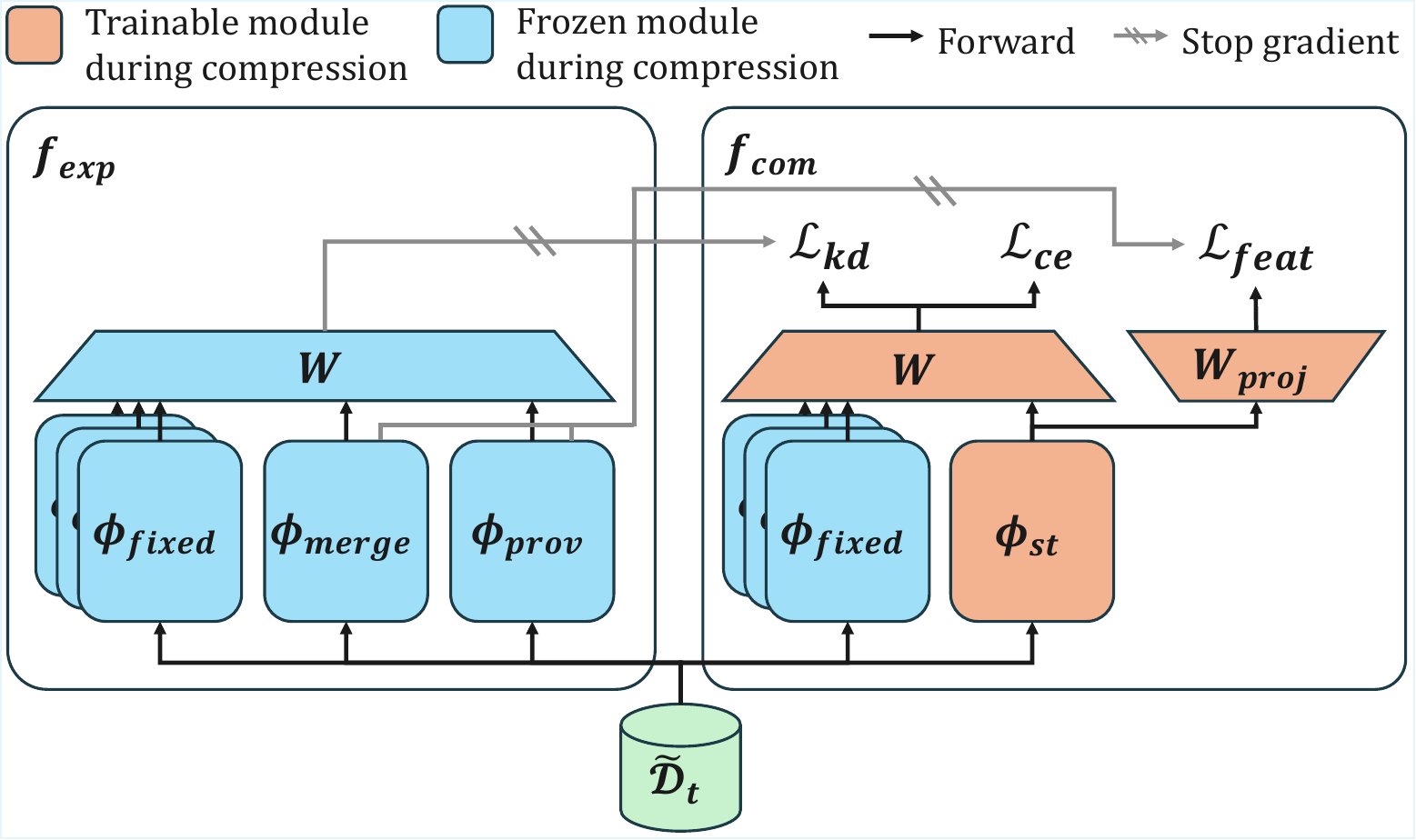}}
    \caption{
      Model consolidation phase of the proposed GRACE framework. The expanded model ($f_{exp}$) acts as a teacher to guide the training of the compressed student model ($f_{com}$). Knowledge transfer is facilitated through a multi-objective loss function: logit-level distillation ($\mathcal{L}_{kd}$), feature-level alignment ($\mathcal{L}_{feat}$), and cross-entropy loss ($\mathcal{L}_{ce}$) to maintain classification performance.
    }
    \label{fig:gracecompression}
  \end{center}
\end{figure}

During the model consolidation phase, the provisional backbone and the mergeable backbone are fused into a single backbone via a knowledge distillation process (see Figure~\ref{fig:gracecompression}). Here, the fully expanded model ($f_{exp}$ with $b$ backbones) serves as the teacher, while a newly initialized network $f_{com}$ with $b - 1$ feature extractors acts as the student. While the student network inherits and freezes the teacher's fixed backbones, its final backbone $\phi_{st}$ remains trainable.

\paragraph{Importance-Aware Student Initialization}

To ensure the student backbone effectively integrates novel features from the provisional backbone while retaining the knowledge of the mergeable backbone, a strategic initialization is required. Simply averaging the weights is suboptimal in incremental settings where class imbalance and varying task sizes introduce significant bias. We propose a strategy that computes a weighting factor $w$ based on the need for knowledge preservation and the potential bias toward new classes.

First, we define the Preservation Factor ($P$), which quantifies the relative importance of the existing knowledge, based on the proportion of previously learned classes and new classes:
\begin{equation}\label{eq:preservationfactor}
    P = \frac{|\mathcal{Y}_{merge}|}{|\mathcal{Y}_{merge}| + |\mathcal{Y}_t|},
\end{equation}
where $|\mathcal{Y}_{merge}|$ denotes the number of classes the mergeable backbone has been trained to represent. A higher $P$ indicates a larger existing knowledge base that requires preservation.

Next, we account for the Bias Factor ($B$). This factor measures the potential training bias toward new classes caused by sample imbalance in $\tilde{\mathcal{D}}_t$. Let $n_{new} = \frac{|\mathcal{D}_t|}{|\mathcal{Y}_t|}$ be the average number of samples per new class, and $n_{old} = \frac{|\mathcal{E}|}{|\tilde{\mathcal{Y}}_{t-1}|}$ be the average number of samples per old class. The bias factor is defined as:
\begin{equation}\label{eq:biasfactor}
    B = \frac{n_{new}}{n_{new} + n_{old}}
\end{equation}

To derive the final weight for the mergeable backbone ($w$), we combine $P$ and $B$ using a $\gamma$-norm power mean. This allows us to control the sensitivity of the initialization to these two factors:
\begin{equation}\label{eq:weightfactor}
    w =\left( \frac{P^\gamma + B^\gamma}{2} \right) ^{\frac{1}{\gamma}}
\end{equation}

Higher $\gamma$ values ensure that if either the preservation requirement or the class bias is high, the initialization leans more heavily toward the mergeable backbone to prevent catastrophic forgetting.

Finally, the student backbone $\phi_{st}$ is initialized as the weighted combination of the mergeable and provisional backbones:
\begin{equation}\label{eq:studentinitialization}
    \phi_{st} = w\phi_{merge} + (1 - w)\phi_{prov}
\end{equation}

By utilizing this weighted approach, the student network starts the optimization process from a point in the weight space that already balances the feature representations of both the old and new tasks, leading to a more stable convergence during the knowledge distillation process.

\begin{table*}[ht]
\centering
\small
\setlength{\tabcolsep}{6pt} 
\caption{\textbf{Experimental results for CIFAR-100 dataset.} The best and second-best accuracy results per scenario are highlighted in \textbf{bold} and \underline{underlined}, respectively.}
\label{tab:cifar100}
\begin{sc}
\begin{tabular}{l c S[table-format=2.2] S[table-format=2.2] c S[table-format=2.2] S[table-format=2.2] c S[table-format=2.2] S[table-format=2.2] c S[table-format=2.2] S[table-format=2.2]}
\toprule
\multirow{2}{*}{Method} & \multicolumn{3}{c}{Base 0 Inc 5} & \multicolumn{3}{c}{Base 0 Inc 10} & \multicolumn{3}{c}{Base 50 Inc 2} & \multicolumn{3}{c}{Base 50 Inc 10} \\
\cmidrule(lr){2-4} \cmidrule(lr){5-7} \cmidrule(lr){8-10} \cmidrule(lr){11-13}
 & {\#P} & {Last} & {Avg} & {\#P} & {Last} & {Avg} & {\#P} & {Last} & {Avg} & {\#P} & {Last} & {Avg} \\
\midrule
Finetune & 0.46 & 4.94 & 17.39 & 0.46 & 8.86 & 26.11 & 0.46 & 1.94 & 5.79 & 0.46 & 8.88 & 22.51 \\
Replay   & 0.46 & 37.25 & 57.00 & 0.46 & 40.05 & 58.65 & 0.46 & 40.12 & 48.93 & 0.46 & 42.11 & 52.09 \\
\midrule
EWC      & 0.46 & 4.90 & 18.73 & 0.46 & 12.41 & 30.82 & 0.46 & 8.50 & 18.37 & 0.46 & 14.41 & 29.32 \\
iCaRL    & 0.46 & 45.34 & 63.08 & 0.46 & 48.64 & 65.06 & 0.46 & 43.45 & 53.35 & 0.46 & 51.90 & 61.04 \\
DER      & 9.20 & {\underline{56.89}} & {\underline{69.31}} & 4.60 & {\underline{60.47}} & {\underline{71.23}} & 11.96 & {\textbf{58.95}} & {\underline{66.09}} & 2.76 & {\underline{62.17}} & {\underline{68.68}} \\
FOSTER   & 0.46 & 47.72 & 61.92 & 0.46 & 54.15 & 66.87 & 0.46 & 50.84 & 60.00 & 0.46 & 58.54 & 66.32 \\
MEMO     & 7.14 & 53.95 & 67.58 & 3.62 & 56.86 & 68.99 & 9.25 & 53.35 & 60.25 & 2.22 & 57.73 & 64.65 \\
BEEF     & 9.20 & 45.32 & 63.78 & 4.60 & 56.86 & 69.59 & 11.96 & 50.62 & 62.29 & 2.76 & 59.86 & 66.83 \\
DGR      & 0.46 & 48.61 & 62.01 & 0.46 & 53.32 & 65.11 & 0.46 & 40.96 & 50.62 & 0.46 & 55.02 & 62.63 \\
\midrule
GRACE    & 4.60 & {\textbf{57.39}} & {\textbf{69.81}} & 2.30 & {\textbf{60.80}} & {\textbf{71.44}} & 3.22 & {\underline{58.09}} & {\textbf{66.42}} & 1.84 & {\textbf{62.55}} & {\textbf{68.85}} \\
\rowcolor[gray]{.95} $\Delta$ & {-50\%} & {+0.50} & {+0.50} & {-50\%} & {+0.33} & {+0.21} & {-73\%} & {-0.86} & {+0.33} & {-33\%} & {+0.38} & {+0.17} \\
\bottomrule
\end{tabular}
\end{sc}
\end{table*}

\paragraph{Dual-Level Knowledge Distillation}

Our model compression phase (Figure~\ref{fig:gracecompression}) leverages logit- and feature-level knowledge distillation to integrate the information from both teacher backbones ($\phi_{merge}$ and $\phi_{prov}$) into the student backbone ($\phi_{st}$).

Let $\mathbf{z}_{exp} = f_{exp}(x)$ denote the logits produced by the fully expanded model, and $\mathbf{z}_{com} = f_{com}(x)$ represent the logits of the student model. The logit-level distillation loss $\mathcal{L}_{kd}$ is defined using the Kullback-Leibler ($KL$) divergence:
\begin{equation}
    \mathcal{L}_{kd} = T^2 \cdot KL \left( \sigma(\mathbf{z}_{exp} / T) \parallel \sigma(\mathbf{z}_{com} / T) \right),
\end{equation}
where $\sigma(\cdot)$ denotes the softmax operator and $T$ is the temperature hyperparameter.

To ensure the student backbone effectively maintains the collective knowledge of both teachers, we employ feature-level distillation. Since the concatenated teacher features $[\phi_{merge}(x), \phi_{prov}(x)]$ reside in a higher-dimensional space ($2d$) than the student features ($d$), we introduce a learnable projection layer $W_{proj}\in\mathbb{R}^{d\times2d}$ to expand the student's feature dimension. The feature distillation loss is then computed via Mean Squared Error ($MSE$):
\begin{equation}
    \mathcal{L}_{feat} = \ell_{MSE}(W_{proj}^\top\phi_{st}, [\phi_{merge}, \phi_{prov}])
\end{equation}

In addition to the distillation terms, we utilize a cross-entropy loss, $\mathcal{L}_{ce} = \ell_{CE}(f_{com}(x), y)$, to ensure the student network maintains high classification accuracy. The final loss $\mathcal{L}_{com}$ is a weighted combination of these components:
\begin{equation}
    \mathcal{L}_{com} = \lambda(\alpha\mathcal{L}_{kd} + \beta\mathcal{L}_{feat}) + (1 - \lambda)\mathcal{L}_{ce},
\end{equation}
where $\lambda$ balances the trade-off between the classification loss and the distillation objective, while $\alpha$ and $\beta$ control the relative contributions of logit-level and feature-level distillation, respectively. Rather than employing a fixed hyperparameter for $\lambda$, we compute its value dynamically at each training stage based on the ratio of old to new classes.

\section{Experiments}\label{sec:experiments}

\subsection{Experimental Settings}

\paragraph{Datasets}
As most CIL image classification benchmarks, based on \cite{rebuffi2017icarl}, we evaluate our model with CIFAR-100 and ImageNet-100. CIFAR-100 \cite{krizhevsky2009cifar100} is a dataset composed of 100 classes, with 500 training and 100 testing images per class, the size of each image being $32\times32$ pixels. ImageNet \cite{russakovsky2015imagenet} is a massive corpus composed of about 1.28 million images across 1000 object classes, and ImageNet-100 is a subset of 100 classes chosen randomly \cite{rebuffi2017icarl}.

In order to maintain an even playing ground across all evaluated methods, we employ the same data augmentation strategy in all our experiments, consisting of random cropping, flipping, and colour jittering. Appendix~\ref{app:augmentation} contains additional evaluations with augmented data, offering a direct performance comparison with TagFex \cite{zheng2025tagfex} and CREATE \cite{chen2025create}.

\paragraph{Protocol}
Following \cite{rebuffi2017icarl}, classes are first shuffled using the NumPy random seed $1993$. Afterward, those classes are divided into incremental tasks adopting the \textit{Base-Increment} setting \cite{zhou2024cilsurvey}, where the \textit{Base} represents the number of classes in the first stage (Base = 0 means that the classes are divided equally), while \textit{Increment} (Inc) denotes the number of classes in each incremental step.

\paragraph{Implementation Details}
Our method is implemented using PyTorch \cite{paszke2017pytorch} and PyCIL \cite{zhou2023pycil}\footnote{The source code for reproducibility is available at \url{https://github.com/ai-digilab/GRACE}.}. We performed our experiments on a workstation equipped with an NVIDIA L40S GPU. All compared methods based on CNN use the same backbone: standard ResNet18 \cite{he2016resnet18} for ImageNet-100 and a modified ResNet32 \cite{rebuffi2017icarl} for CIFAR-100. The exemplar buffer size is set to 2,000 images for both CIFAR-100 and ImageNet-100, and we use \textit{herding} \cite{rebuffi2017icarl} as the strategy to select the exemplars.

\begin{table*}[ht]
\centering
\small
\setlength{\tabcolsep}{6pt}
\caption{\textbf{Experimental results for ImageNet-100 dataset.} The best and second-best accuracy results per scenario are highlighted in \textbf{bold} and \underline{underlined}, respectively. Results for BEEF are omitted for the \textit{Base 0 Inc 5} and \textit{Base 50 Inc 10} settings, as the model's memory requirements exceeded our 48GB GPU capacity.}
\label{tab:imagenet100}
\begin{sc}
\begin{tabular}{l c S[table-format=2.2] S[table-format=2.2] c S[table-format=2.2] S[table-format=2.2] c S[table-format=2.2] S[table-format=2.2]}
\toprule
\multirow{2}{*}{Method} & \multicolumn{3}{c}{Base 0 Inc 5} & \multicolumn{3}{c}{Base 0 Inc 10} & \multicolumn{3}{c}{Base 50 Inc 10} \\
\cmidrule(lr){2-4} \cmidrule(lr){5-7} \cmidrule(lr){8-10}
 & {\#P} & {Last} & {Avg} & {\#P} & {Last} & {Avg} & {\#P} & {Last} & {Avg} \\
\midrule
Finetune & 11.17 & 4.66 & 16.82 & 11.17 & 9.28 & 26.28 & 11.17 & 9.58 & 24.53 \\
Replay   & 11.17 & 36.34 & 55.62 & 11.17 & 42.64 & 60.26 & 11.17 & 41.46 & 51.86 \\
\midrule
EWC      & 11.17 & 5.60 & 18.83 & 11.17 & 10.86 & 28.06 & 11.17 & 11.68 & 26.35 \\
iCaRL    & 11.17 & 44.48 & 61.86 & 11.17 & 50.46 & 67.23 & 11.17 & 53.96 & 63.42 \\
DER      & 223.4 & {{\ul 64.62}} & {\textbf{73.66}} & 111.7 & {\textbf{67.00}} & {\textbf{76.16}} & 67.02 & {{\ul 73.34}} & {\textbf{79.59}} \\
FOSTER   & 11.17 & 51.38 & 63.07 & 11.17 & 60.78 & 69.22 & 11.17 & 67.40 & 74.22 \\
MEMO     & 170.6 & 56.02 & 68.38 & 86.72 & 61.04 & 72.05 & 53.14 & 60.16 & 67.56 \\
BEEF     & {223.4} & {--} & {--} & 111.7 & 65.60 & 74.71 & {67.02} & {--} & {--} \\
DGR      & 11.17 & 48.68 & 60.87 & 11.17 & 54.18 & 63.16 & 11.17 & 65.15 & 72.21 \\
\midrule
GRACE    & 134.04 & {\textbf{64.86}} & {{\ul 73.52}} & 78.19 & {{\ul 66.76}} & {{\ul 76.06}} & 44.68 & {\textbf{73.50}} & {{\ul 79.55}} \\
\rowcolor[gray]{.95} $\Delta$ & {-40\%} & {+0.24} & {-0.14} & {-30\%} & {-0.24} & {-0.10} & {-33\%} & {+0.16} & {-0.04} \\
\bottomrule
\end{tabular}
\end{sc}
\end{table*}
    
\subsection{Benchmark Results}

We thoroughly evaluate our method in a variety of incremental protocols. For each scenario, we report the final model size in millions of parameters (\textit{\#P}), the test accuracy upon completion of all incremental steps (\textit{Last}), and the average test accuracy computed across all stages (\textit{Avg}). Detailed numerical results for the CIFAR-100 and ImageNet-100 datasets are provided in Table~\ref{tab:cifar100} and Table~\ref{tab:imagenet100}, respectively.

\noindent\textbf{GRACE achieves state-of-the-art performance with up to a 73\% reduction in parameter cost}. Compared to top-performing methods in terms of accuracy, GRACE achieves competitive accuracy (with marginal trade-offs depending on the incremental setting) while substantially reducing backbone expansion through adaptive compression. This advantage is especially pronounced in high-task-count protocols, as more incremental steps imply more redundant parameters, as well as more opportunities to optimize the architecture and avoid unnecessary parameter expansion. Under the \textit{Base 50 Inc 10} setting (6 tasks), GRACE achieves a 33\% reduction in parameters; however, in the more granular \textit{Base 50 Inc 2} setting (26 tasks), this efficiency peaks at a 73\% reduction in the total number of parameters. These results demonstrate that GRACE's adaptive compression mechanism is most effective in long-sequenced scenarios.

Conversely, the disadvantage of other fixed-capacity methods becomes most evident in these same settings: while completely avoiding parameter expansion, numerous updates on static architectures leads to larger performance gaps with GRACE. For instance, the accuracy margin over FOSTER (as representative of the fixed-capacity paradigm) more than doubles as the number of updates increases, rising from 6 points in ImageNet-100 \textit{Base 50 Inc 10} to more than 13 points in the \textit{Base 0 Inc 5} configuration, which highlights the limitations of non-expansionist networks and the importance of strategic network expansion.

\begin{table}[ht]
\centering
\small
\setlength{\tabcolsep}{4pt}
\caption{\textbf{Memory-aligned comparison results.} Comparison on CIFAR-100 and ImageNet-100 under identical memory constraints. Differences in parameter count are reallocated to the rehearsal memory size, M, to ensure an equitable evaluation.}
\label{tab:memoryaligned}
\begin{sc}
\begin{tabular}{l S[table-format=5.0] S[table-format=2.2] S[table-format=2.2] S[table-format=5.0] S[table-format=2.2] S[table-format=2.2]}
\toprule
\multirow{3}{*}{Method} & \multicolumn{6}{c}{CIFAR-100} \\
\cmidrule(lr){2-7}
 & \multicolumn{3}{c}{Base 0 Inc 5} & \multicolumn{3}{c}{Base 0 Inc 10} \\
\cmidrule(lr){2-4} \cmidrule(lr){5-7}
 & {M} & {Last} & {Avg} & {M} & {Last} & {Avg} \\
\midrule
Replay & 13466 & 60.83 & 73.66 & 7431 & 56.37 & 70.26 \\
iCaRL  & 13466 & 61.40 & 74.02 & 7431 & 59.08 & 71.41 \\
DER    & 2000  & 56.89 & 69.31 & 2000 & 60.47 & 71.23 \\
FOSTER & 13466 & {{\ul 63.28}} & {{\ul 74.94}} & 7431 & 60.26 & {{\ul 72.56}} \\
MEMO   & 4771  & 62.14 & 72.96 & 3312 & {{\ul 61.57}} & 72.01 \\
BEEF   & 2000  & 45.32 & 63.78 & 2000 & 56.86 & 69.59 \\
DGR    & 13466 & 56.49 & 65.76 & 7431 & 57.37 & 65.46 \\
GRACE  & 8035  & {\textbf{65.17}} & {\textbf{75.76}} & 5017 & {\textbf{65.28}} & {\textbf{74.63}} \\
\rowcolor[gray]{.95} $\Delta$ & {--} & {+1.89} & {+0.82} & {--} & {+3.71} & {+2.07} \\
\midrule
\multirow{3}{*}{Method} & \multicolumn{6}{c}{ImageNet-100} \\
\cmidrule(lr){2-7}
 & \multicolumn{3}{c}{Base 0 Inc 5} & \multicolumn{3}{c}{Base 0 Inc 10} \\
\cmidrule(lr){2-4} \cmidrule(lr){5-7}
 & {M} & {Last} & {Avg} & {M} & {Last} & {Avg} \\
\midrule
Replay & 7639 & 54.96 & 70.45 & 4671 & 53.58 & 68.10 \\
iCaRL  & 7639 & 59.86 & 73.10 & 4671 & 59.42 & 71.99 \\
DER    & 2000 & {{\ul 64.62}} & {{\ul 73.66}} & 2000 & {{\ul 67.00}} & {{\ul 76.16}} \\
FOSTER & 7639 & 59.82 & 72.95 & 4671 & 61.16 & 71.98 \\
MEMO   & 3403 & 60.60 & 72.46 & 2664 & 65.14 & 73.64 \\
BEEF   & 2000 & {--}  & {--}  & 2000 & 65.60 & 74.71 \\
DGR    & 7639 & 60.26 & 68.68 & 4671 & 58.96 & 66.36 \\
GRACE  & 4374 & {\textbf{68.84}} & {\textbf{77.28}} & 2890 & {\textbf{69.22}} & {\textbf{77.32}} \\
\rowcolor[gray]{.95} $\Delta$ & {--} & {+4.22} & {+3.62} & {--} & {+2.22} & {+1.16} \\
\bottomrule
\end{tabular}
\end{sc}
\end{table}

\subsection{Memory-Aligned Evaluation}

As proposed in \cite{zhou2022memo, zhou2024cilsurvey}, we conducted a memory-aligned comparison. In this setup, disparities in model parameters and expansion rates are offset by adjusting the exemplar memory size, ensuring all methods operate under a uniform memory budget. This budget is capped by the most memory-expensive approaches, specifically DER and BEEF. Other approaches are granted additional rehearsal memory beyond the baseline of 2000 exemplars. Given that each saved backbone (ResNet32) comprises 463,504 floating-point parameters, while a CIFAR image consists of $32\times32\times3$ bytes, a single backbone represents a memory footprint equivalent to approximately 603 exemplars. In the case of ImageNet, a single ResNet18 backbone (11,176,512 parameters) represents a memory footprint comparable to 297 images ($224\times224\times3$ bytes).

\noindent\textbf{GRACE surpasses state-of-the-art methods under equivalent memory constraints}. The results presented in Table~\ref{tab:memoryaligned} demonstrate that GRACE outperforms the current state-of-the-art, with improvements of up to 4.22 percentage points under identical memory budgets. This experimental setup demonstrates the efficiency of strategic network expansion coupled with effective backbone consolidation, effectively converting memory savings into accuracy gains.

\begin{figure}[ht]
    \centering
    \begin{subfigure}[b]{0.5\columnwidth}
        \centering
        \includegraphics[width=\textwidth]{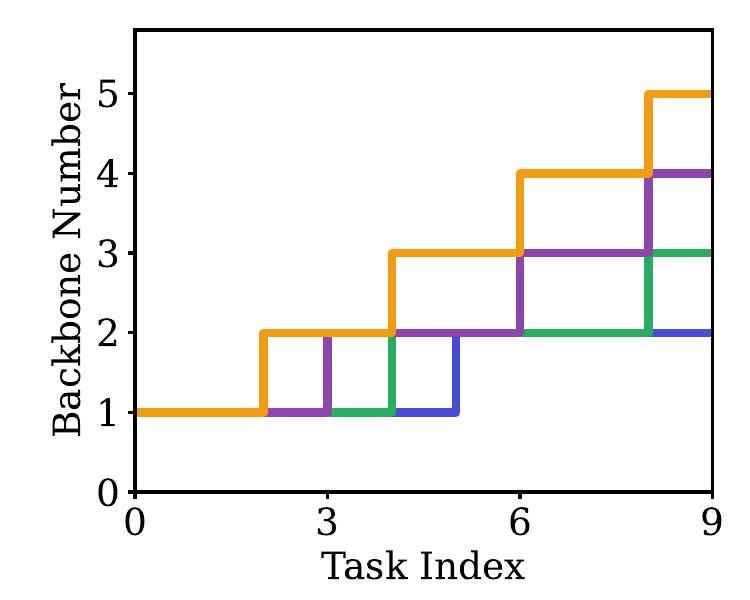}
    \end{subfigure}%
    \hfill
    \begin{subfigure}[b]{0.5\columnwidth}
        \centering
        \includegraphics[width=\textwidth]{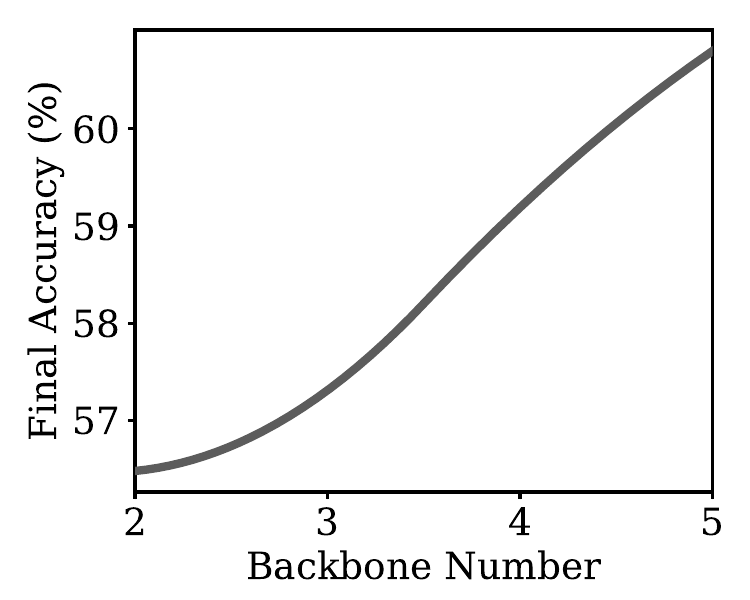}
    \end{subfigure}

    \caption{GRACE's expansion dynamics (left) and accuracy change (right) on CIFAR-100 \textit{Base 0 Inc 10} with different threshold and decay configurations.}
    \label{fig:versatility}
\end{figure}

\subsection{Analysis of Model Adaptability}

We evaluated the versatility of our method on CIFAR-100 across a spectrum of threshold and decay configurations (detailed in Figure~\ref{fig:versatility} and Appendix~\ref{appendix:grid_search}). The results demonstrate that our approach offers a control mechanism over the expansion rate and therefore terminal model size. By tuning these hyperparameters, the model can be tailored to meet the specific constraints of diverse deployment scenarios, ranging from resource-constrained edge devices to high-performance servers.

\subsection{Ablation Study}

To evaluate the individual contributions of each component within the GRACE compression phase, we conducted an ablation study across three key elements: Importance-Aware Student Initialization (\textit{Init.}), feature-level knowledge distillation (\textit{Feat KD}), and standard logit distillation (\textit{Logit KD}). By testing all possible combinations, we isolated their respective effects; as shown in Table~\ref{tab:ablation}, the full GRACE configuration achieves the highest accuracy, confirming that each module contributes positively to the final result.

Apart from the overall efficacy of the full framework, the ablation results highlight the critical role of student initialization. Consequently, we compared the standard weight-average technique with our proposed importance-aware method. As shown in Table~\ref{tab:studentini}, our approach consistently outperforms the baseline across all evaluated CIFAR-100 benchmarks, achieving improvements of more than 2 percentage points.

\begin{table}[t]
\centering
\small
\setlength{\tabcolsep}{0pt}
\caption{\textbf{Ablation results on GRACE compression.} Comparison of different compression components in CIFAR-100 under the \textit{Base 0 Inc 10} setting.}
\label{tab:ablation}
\begin{sc}
    \begin{tabularx}{\columnwidth}{@{\extracolsep{\fill}}l ccc cc@{}}
    \toprule
    \multirow{2}{*}{Method} & \multicolumn{3}{c}{Components} & \multirow{2}{*}{Last} &  \multirow{2}{*}{Avg}\\
    \cmidrule(lr){2-4}
    & Init. & Feat KD & Logit KD & & \\
    \midrule
    DER (Base) & -- & -- & -- & 60.47 & 71.23 \\
    \midrule
    \multirow{7}{*}{GRACE} 
    & \ding{55} & \ding{55} & \ding{51} & 38.41 & 58.23 \\
    & \ding{55} & \ding{51} & \ding{55} & 45.36 & 60.03 \\
    & \ding{51} & \ding{55} & \ding{55} & 54.11 & 68.69 \\
    & \ding{55} & \ding{51} & \ding{51} & 47.64 & 61.68 \\
    & \ding{51} & \ding{55} & \ding{51} & 58.90 & 70.28 \\
    & \ding{51} & \ding{51} & \ding{55} & 59.54 & 71.17 \\
    & \ding{51} & \ding{51} & \ding{51} & \textbf{60.80} & \textbf{71.44} \\
    \bottomrule
    \end{tabularx}
\end{sc}
\end{table}

\begin{table}[t]
\centering
\caption{\textbf{Impact of Importance-Aware Student Initialization.} Evaluation across four CIFAR-100 benchmarks (\textit{Base 0 Inc 5/10} and \textit{Base 50 Inc 2/10}). Results contrast standard average initialization with our proposed importance-aware technique.}
\label{tab:studentini}
\small
\begin{sc}    
\begin{tabular}{l S[table-format=2.2] S[table-format=2.2] S[table-format=2.2] S[table-format=2.2]}
\toprule
\multirow{3}{*}{Method} & \multicolumn{4}{c}{CIFAR-100 Base 0} \\
\cmidrule(lr){2-5}
 & \multicolumn{2}{c}{Inc 5} & \multicolumn{2}{c}{Inc 10} \\
\cmidrule(lr){2-3} \cmidrule(lr){4-5}
 & {Last} & {Avg} & {Last} & {Avg} \\
\midrule
Avg. Baseline & 56.99 & 69.50 & 60.02 & 70.64 \\
Imp.-Aw. (Ours) & \textbf{57.39} & \textbf{69.81} & \textbf{60.80} & \textbf{71.44} \\
\rowcolor[gray]{.95} $\Delta$ & {+0.40} & {+0.31} & {+0.78} & {+0.80} \\
\midrule
\multirow{3}{*}{Method} & \multicolumn{4}{c}{CIFAR-100 Base 50} \\
\cmidrule(lr){2-5}
 & \multicolumn{2}{c}{Inc 2} & \multicolumn{2}{c}{Inc 10} \\
\cmidrule(lr){2-3} \cmidrule(lr){4-5}
 & {Last} & {Avg} & {Last} & {Avg} \\
\midrule
Avg. Baseline & 56.38 & 65.58 & 60.29 & 67.58 \\
Imp.-Aw. (Ours) & \textbf{58.09} & \textbf{66.42} & \textbf{62.55} & \textbf{68.85} \\
\rowcolor[gray]{.95} $\Delta$ & {+1.71} & {+0.84} & {+2.26} & {+1.27} \\
\bottomrule
\end{tabular}
\end{sc}
\end{table}

\section{Conclusions}\label{sec:conclusion}

In this paper we presented GRACE, a capacity-aware backbone scaling framework for Class Incremental Learning. By replacing arbitrary expansion protocols with an informed decision-making mechanism, GRACE selectively expands its architecture only when necessary. This selective growth is coupled with an enhanced compression phase designed to mitigate performance degradation during model consolidation. Extensive evaluation on CIFAR-100 and ImageNet confirms that GRACE successfully balances high predictive performance with superior memory efficiency. Beyond achieving a memory footprint reduction of up to 73\% across standard benchmarks, GRACE surpasses current state-of-the-art by more than 4 percentage points under identical memory constraints. These results suggest that informed architectural growth is a viable path toward sustainable, long-term incremental learning.

\section*{Acknowledgements}

Special thanks to Ibai Roman Txopitea for his critical reviews and insightful suggestions.


\bibliography{example_paper}
\bibliographystyle{icml2026}

\newpage
\appendix
\onecolumn

\section{Formalization of the Grow-Assess-Compress Framework}\label{app:gracealgorithm}

Algorithm~\ref{alg:grace} details the incremental training workflow for each task $t$. It illustrates the operational logic of the proposed grow-assess-compress cycle, providing a formal mechanism for balancing novel feature acquisition with the preservation of legacy knowledge.

\begin{algorithm}[H]
    \caption{GRACE Incremental Training Procedure}
    \label{alg:grace}
    \begin{algorithmic}[1]
        \STATE {\bfseries Input:} Task dataset $\tilde{\mathcal{D}}_t$, current threshold $\tau_t$, threshold decay $\rho$
        \STATE {\bfseries Output:} Updated $\phi_{merge}$, $\phi_{fixed}$, $\tau_{t+1}$
        
        \STATE // Stage I: Grow
        \STATE Initialize provisional backbone $\phi_{prov}$ and expand classifier $W$
        \STATE Train $\phi_{prov}$ with $\tilde{\mathcal{D}}_t$ // Other backbones remain frozen
        
        \STATE // Stage II: Assess
        \STATE $\tilde{eRank} \leftarrow$ Calculate saturation of $\phi_{merge}$ using $\tilde{\mathcal{D}}_t$
        
        \IF{$\tilde{eRank} < \tau_t$}
            \STATE // Stage III: Compress (Consolidation Case)
            \STATE $\phi_{st} \leftarrow$ Merge $\phi_{prov}$ and $\phi_{merge}$ using $\tilde{\mathcal{D}}_t$
            \STATE $\phi_{merge} \leftarrow \phi_{st}$
            \STATE $\tau_{t+1} = \rho \cdot \tau_t$ // Decay threshold
        \ELSE
            \STATE // Expansion Case
            \STATE Move $\phi_{merge}$ to fixed knowledge base $\phi_{fixed}$
            \STATE $\phi_{merge} \leftarrow \phi_{prov}$ // Promote provisional to mergeable
            \STATE $\tau_{t+1} = \tau_1$ // Reset threshold for new backbone
        \ENDIF
    \end{algorithmic}
\end{algorithm}

\section{Effective Rank Calculation}
\label{appendix:erank}

This section details the calculation of the effective-rank-based saturation score, $\tilde{eRank}$. First, we construct the matrix $Z \in \mathbb{R}^{|\tilde{\mathcal{D}}_t| \times d}$ by stacking the features extracted by $\phi_{\text{merge}}$ for every data sample in $\tilde{\mathcal{D}}_t$. We then center the data to obtain $\tilde{Z} = Z - \bar{Z}$ and perform Singular Value Decomposition (SVD) to obtain the singular values, $\sigma$. These singular values are normalized to form a probability distribution $p$, where each element $p_i$ is defined as:
\begin{equation}
    p_i = \frac{\sigma_i}{\sum_{j=1}^k\sigma_j}
\end{equation}

The $eRank$ is then calculated as the exponential of the Shannon Entropy of this distribution:
\begin{equation}
    eRank = \exp\left( -\sum_{i=1}^k p_i \ln{p_i} \right)
\end{equation}

Finally, we normalize the $eRank$ to derive a saturation score $\tilde{eRank} \in [0, 1]$:
\begin{equation}
    \tilde{eRank} = \frac{eRank}{\min(|\tilde{\mathcal{D}}_t|, d)}
\end{equation}

\section{Saturation Measure Comparison}\label{appendix:erankcomparison}

\begin{figure}[H]
    \centering

    \begin{subfigure}[b]{0.32\textwidth}
        \centering
        \includegraphics[width=\textwidth]{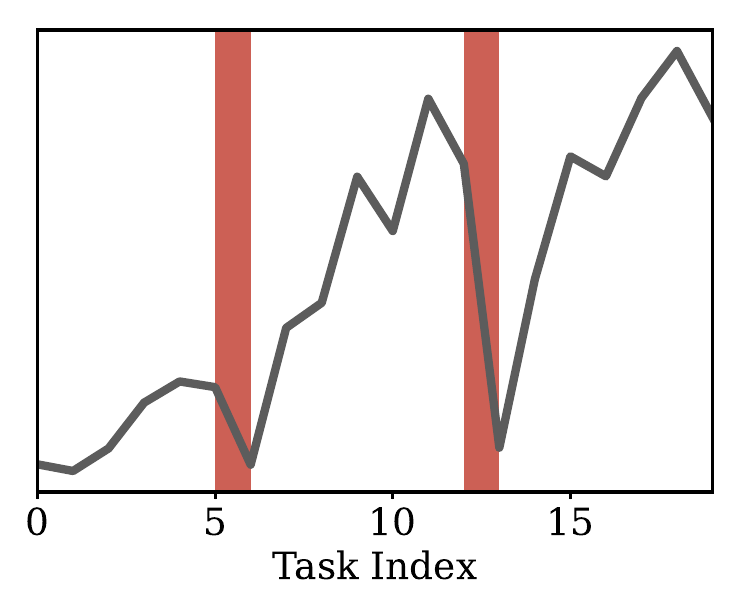}
        \caption{Mean Fisher Information}
    \end{subfigure}%
    \hfill
    \begin{subfigure}[b]{0.32\textwidth}
        \centering
        \includegraphics[width=\textwidth]{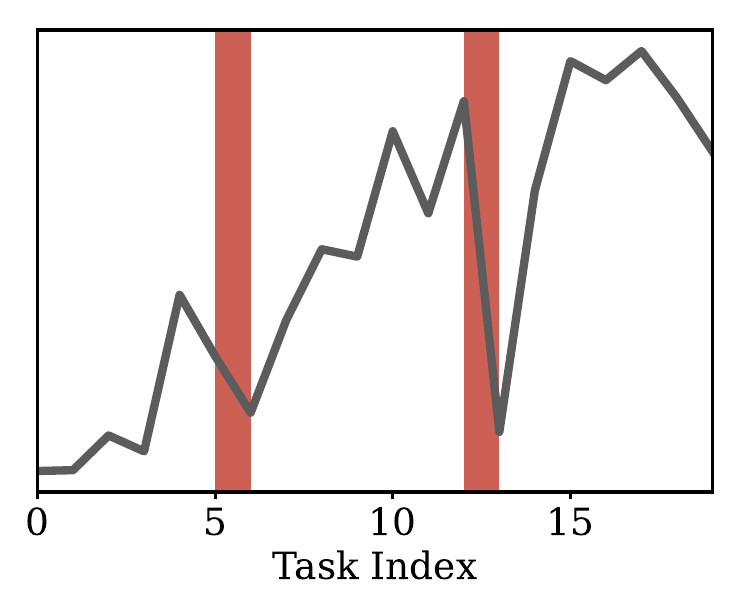}
        \caption{Adversarial Local Sharpness}
    \end{subfigure}%
    \hfill
    \begin{subfigure}[b]{0.32\textwidth}
        \centering
        \includegraphics[width=\textwidth]{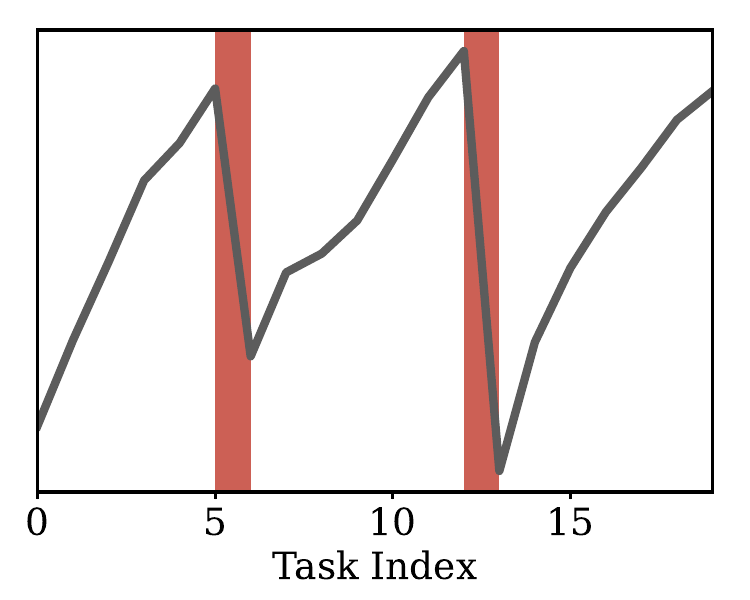}
        \caption{Effective Rank}
    \end{subfigure}
    
    \begin{subfigure}[b]{0.32\textwidth}
        \centering
        \includegraphics[width=\textwidth]{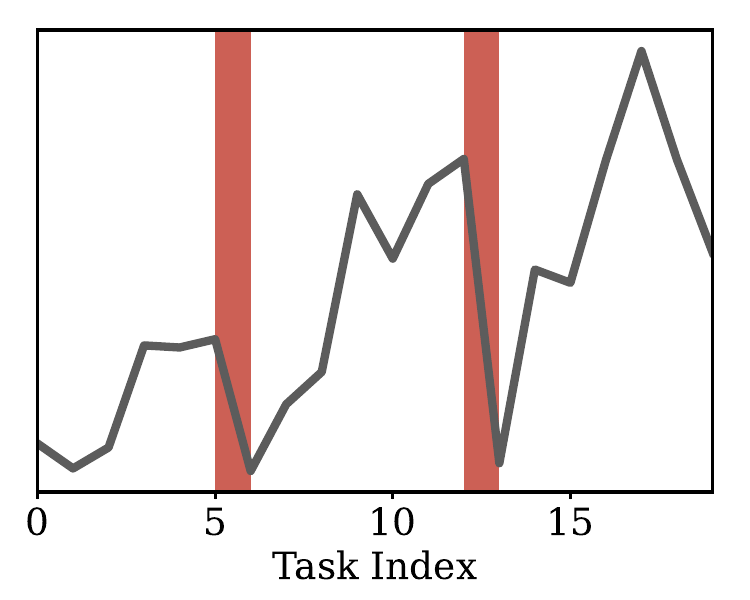}
        \caption{Hessian Spectral Radius}
    \end{subfigure}%
    \hfill
    \begin{subfigure}[b]{0.32\textwidth}
        \centering
        \includegraphics[width=\textwidth]{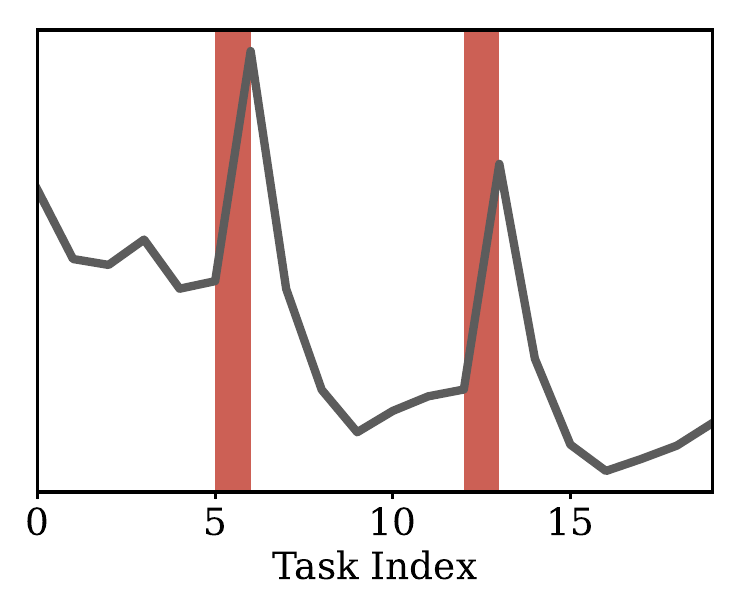}
        \caption{Weight Norm}
    \end{subfigure}%
    \hfill
    \begin{subfigure}[b]{0.32\textwidth}
        \centering
        \includegraphics[width=\textwidth]{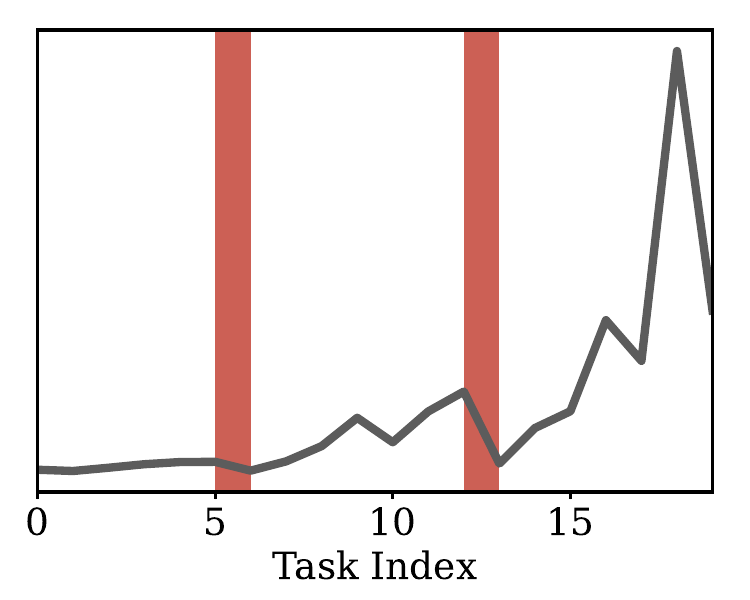}
        \caption{Expected Local Sharpness}
    \end{subfigure}%

    \caption{Comparison of different saturation measures over CIFAR-100 \textit{Base 0 Inc 5}, with fixed expansions in tasks 6 and 13.}
    \label{fig:saturationcomparison}
\end{figure}

We compared six saturation measures to track model capacity evolution using a CIFAR-100 \textit{Base 0 Inc 5} setup, with fixed expansion steps at tasks 6 and 13. Figure~\ref{fig:saturationcomparison} displays the saturation levels of the mergeable backbone calculated after each training stage. Effective Rank demonstrates a nearly symmetric exhaustion pattern across the three learning ranges (indicated by red vertical dividers), where the same mergeable backbone acquires new knowledge incrementally. Other measures, likely due to the interference of secondary training factors, do not exhibit the expected monolithic increase.

\section{Threshold Sensitivity and Grid Search}
\label{appendix:grid_search}

\begin{figure}[ht]
    \centering

    \begin{subfigure}[b]{0.7\textwidth}
        \centering
        \includegraphics[width=\textwidth]{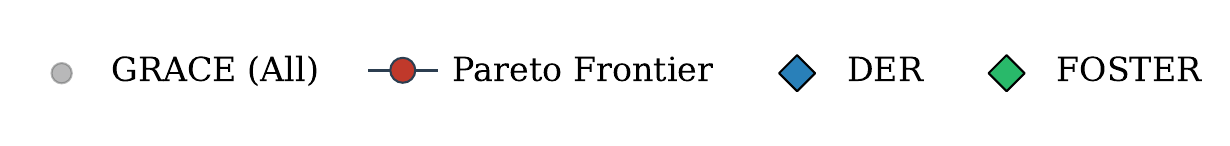}
    \end{subfigure}
    
    \begin{subfigure}[b]{0.24\textwidth}
        \centering
        \includegraphics[width=\textwidth]{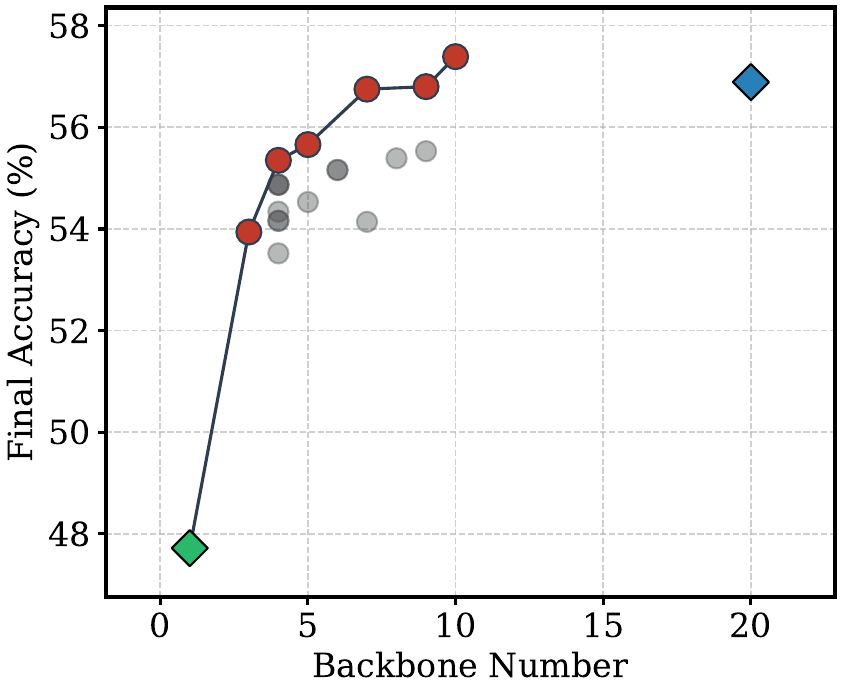}
        \caption{CIFAR-100 B0 Inc5}
    \end{subfigure}%
    \hfill
    \begin{subfigure}[b]{0.24\textwidth}
        \centering
        \includegraphics[width=\textwidth]{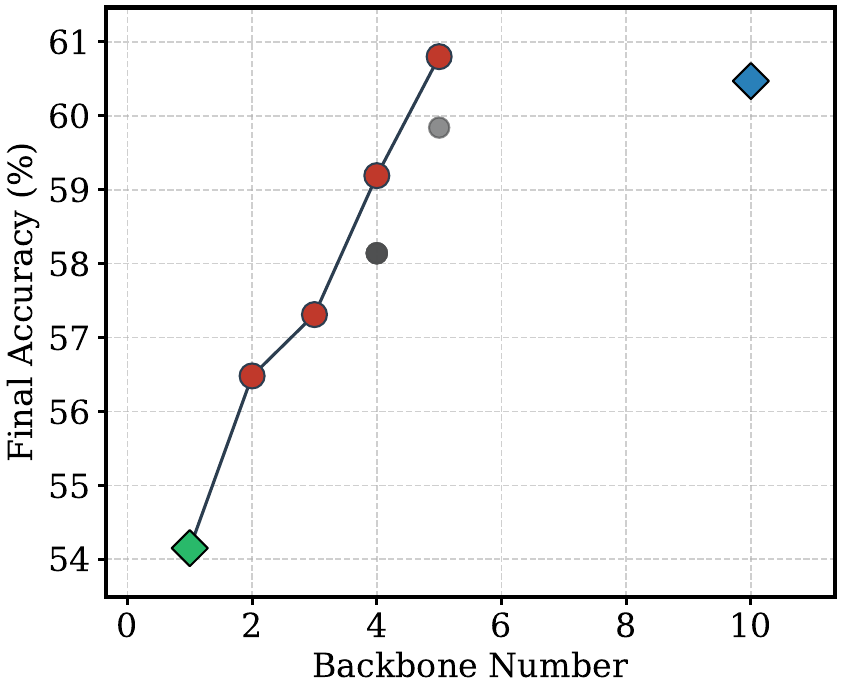}
        \caption{CIFAR-100 B0 Inc10}
    \end{subfigure}%
    \hfill
    \begin{subfigure}[b]{0.24\textwidth}
        \centering
        \includegraphics[width=\textwidth]{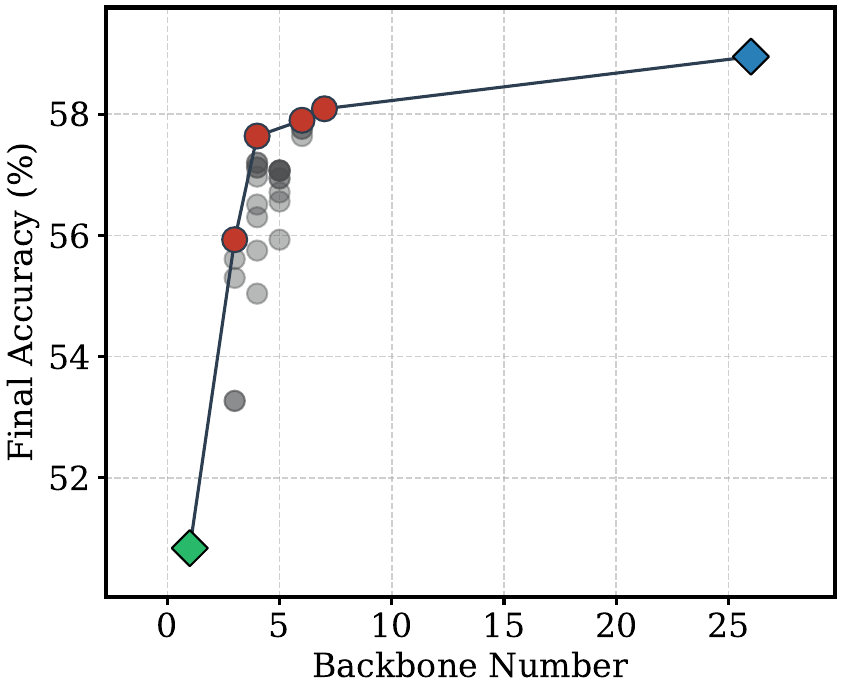}
        \caption{CIFAR-100 B50 Inc2}
    \end{subfigure}%
    \hfill
    \begin{subfigure}[b]{0.24\textwidth}
        \centering
        \includegraphics[width=\textwidth]{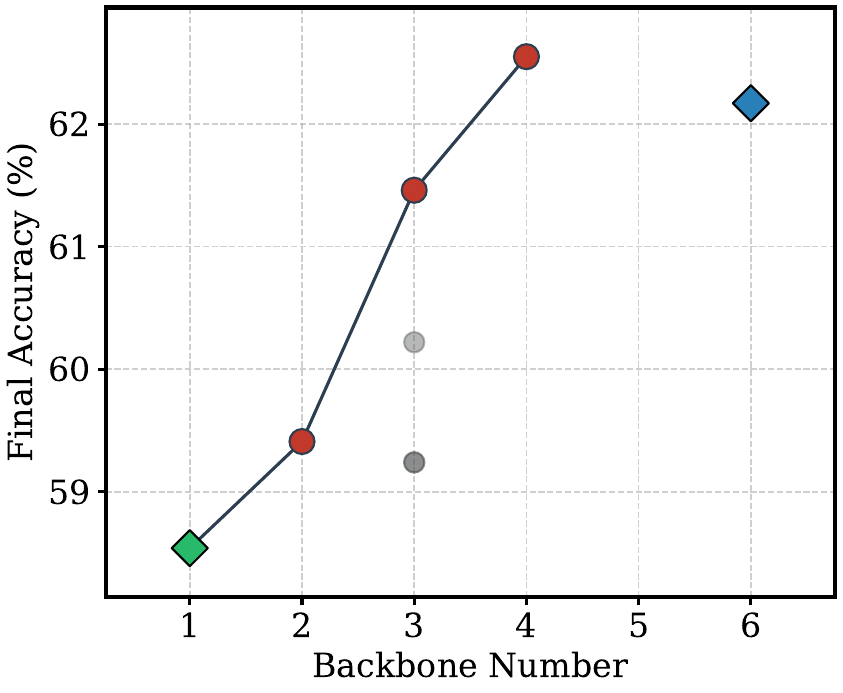}
        \caption{CIFAR-100 B50 Inc10}
    \end{subfigure}

    \caption{Sensitivity analysis of threshold and threshold decay values on CIFAR-100. The plots compare performance across different protocol settings, highlighting the Pareto frontier for hyperparameter selection. Note that some points may be overlapped, meaning that runs with different threshold values produced equal compression decisions.}
    \label{fig:paretocomparison}
\end{figure}

To identify the optimal configuration for our model, we performed an extensive grid search over the threshold and threshold decay parameters, with values $\tau_1 \in [0.5, 0.8]$ and $\rho \in [0.9, 0.98]$. These experiments were conducted using the CIFAR-100 dataset under different protocols. Figure~\ref{fig:paretocomparison} illustrates the trade-offs between these settings, providing a Pareto-style comparison of performance and model size (expressed by backbone number).

\section{Experiments with Augmented Data}\label{app:augmentation}

This section evaluates performance on CIFAR-100 using the AutoAugment \cite{cubuk2019autoaugment} strategy. Here, we benchmark our method against DER \cite{yan2021der} and specific augmentation-dependent approaches, \textit{i.e.} TagFex \cite{zheng2025tagfex} and CREATE \cite{chen2025create}. These were omitted from the primary comparisons to ensure a consistent experimental baseline, as they rely on specialized augmentation protocols.

\begin{table}[H]
\centering
\small
\setlength{\tabcolsep}{6pt}
\caption{\textbf{CIFAR-100 results with AutoAugment.} Comparative analysis for \textit{Base 0 Inc 10} and \textit{Base 50 Inc 10} incremental settings. The best and second-best accuracy results per scenario are highlighted in \textbf{bold} and \underline{underlined}, respectively.}
\label{tab:cifar100aa}
\begin{sc}
\begin{tabular}{l S[table-format=1.2] S[table-format=2.2] S[table-format=2.2] S[table-format=1.2] S[table-format=2.2] S[table-format=2.2]}
\toprule
\multirow{2}{*}{Method} & \multicolumn{3}{c}{Base 0 Inc 10} & \multicolumn{3}{c}{Base 50 Inc 10} \\
\cmidrule(lr){2-4} \cmidrule(lr){5-7}
 & {\#P} & {Last} & {Avg} & {\#P} & {Last} & {Avg} \\
\midrule
DER     & 4.60 & {{\ul 68.77}} & {\textbf{77.12}} & 2.76 & 69.07 & 74.07 \\
TagFex  & 4.60 & {\textbf{68.86}} & {{\ul 77.05}} & 2.76 & {{\ul 69.28}} & {\textbf{74.42}} \\
CREATE  & 0.87 & 55.64 & 69.88 & 0.87 & 57.86 & 67.74 \\
GRACE   & 2.30 & 67.98 & 77.00 & 1.84 & {\textbf{69.42}} & {{\ul 74.10}} \\
\bottomrule
\end{tabular}
\end{sc}
\end{table}

The results in Table~\ref{tab:cifar100aa} demonstrate that on the augmented CIFAR-100 dataset, our method remains consistent with our primary findings, achieving performance comparable to state-of-the-art approaches while maintaining a reduced memory budget.


\end{document}